\documentclass[10pt,twocolumn,letterpaper]{article}

 \usepackage{cvpr}              

\usepackage{multirow}
\usepackage{bm}

\usepackage[ruled,vlined]{algorithm2e}

\definecolor{cvprblue}{rgb}{0.21,0.49,0.74}
\usepackage[pagebackref,breaklinks,colorlinks,allcolors=cvprblue]{hyperref}
\usepackage{graphicx}
\usepackage{booktabs}
\usepackage{makecell}
\usepackage[ruled,vlined]{algorithm2e}
\usepackage[table]{xcolor}
\definecolor{deepgreen}{RGB}{0,100,0}
\usepackage[percent]{overpic}
\usepackage{tabularx}
\usepackage{float}
\usepackage{hyperref}
\def\paperID{27} 
\def\confName{CVPR}
\def\confYear{2026}

\title{Omni-NegCLIP: Enhancing CLIP with Front-Layer Contrastive Fine-Tuning for Comprehensive Negation Understanding}

\author{
Jingqi Xu\\
University of Southern California\\
Los Angeles, CA, USA\\
{\tt\small jingqixu@usc.edu}
}

\begin{document}
\maketitle
\begin{abstract}
Vision-Language Models (VLMs) have demonstrated strong capabilities across a wide range of multimodal tasks. However, recent studies have shown that VLMs, such as CLIP, perform poorly in understanding negation expressions, which are common in natural language. 
In this work, we propose \textbf{Omni-NegCLIP}, a fine-tuned CLIP model that improves CLIP's understanding of two types of negation, namely \textit{presence-based negation} and \textit{absence-based negation}, which correspond to negated expressions of objects that are actually present in an image and those that may plausibly exist in an image but are in fact absent, respectively, by modifying CLIP's original InfoNCE contrastive loss. Specifically, we design a \textit{presence-based contrastive objective} that pulls image embeddings closer to their original caption embeddings while pushing them away from the corresponding presence-based negated caption embeddings, and an \textit{absence-based contrastive objective} that aligns image embeddings with both original and absence-based negated caption embeddings while maintaining a semantic distinction between the two text embeddings. Based on our observation that the front transformer layers of CLIP’s text encoder have stronger learning ability for negated text than the later layers, we fine-tune the front transformer layers of the CLIP text encoder at each training step using the combined contrastive objective. Experimental results show that, compared with pretrained CLIP, \textbf{Omni-NegCLIP} improves performance on presence-based negation and absence-based negation tasks by up to 52.65\% and 12.50\%, respectively, without sacrificing general capability in image-text retrieval and even improving it by up to 19.62\%. Compared with prior works, \textbf{Omni-NegCLIP} demonstrates a more comprehensive ability to understand multiple types of negation tasks. Code is available at \url{https://github.com/Jingqi-Xu/Omni-NegCLIP}.
\end{abstract}    
\section{Introduction}
\label{sec:intro}
Recent vision-language models (VLMs)~\cite{li2022blip,alayrac2022flamingo,radford2021learning,singh2022flava,wang2022omnivl,zhai2022lit,jia2021scaling,li2021align,yuan2021florence} have demonstrated strong capabilities across a wide range of multimodal tasks~\cite{liu2024mmbench,yue2024mmmu,choi2022perception,flux2024} by jointly leveraging visual and textual information. Among them, CLIP~\cite{radford2021learning} is one of the most representative and influential models. Through contrastive learning, CLIP learns a shared multimodal embedding space in which semantically aligned image-caption pairs are pulled closer together, while semantically mismatched pairs are pushed farther apart. Owing to this powerful image-text alignment ability, CLIP has achieved strong performance on tasks such as zero-shot image classification, semantic retrieval~\cite{hendriksen2024assessing}, and image-text matching~\cite{zhang2024statistical}. More importantly, CLIP also serves as a fundamental backbone for many downstream models and systems~\cite{liu2023visual,li2024llava,awadalla2023openflamingo,li2022grounded}. Therefore, its capability of accurately understanding image-text semantics plays a crucial role in the performance of a broad range of vision-language applications.

In recent years, several studies~\cite{hu2025decoupled,patel2024tripletclip,li2025enhancing,huang2024llm2clip} have been conducted to improve CLIP, especially its ability to understand structured and compositionally complex sentences. However, recent findings~\cite{singh2024learn,park2025know,wang2022learn} suggest that CLIP struggles with negation understanding. In particular, it exhibits low sensitivity to negation cues in captions, such as "no," "not," and "without". Since negation is a pervasive phenomenon in natural language~\cite{khemlani2014negations}, this weakness limits CLIP’s commonsense reasoning ability in real-world settings and reduces its robustness in downstream tasks such as image-text retrieval and text-to-image generation. Despite the importance of this problem, training strategies specifically designed to improve CLIP’s understanding of negation remain largely underexplored. Only a few studies~\cite{singh2024learn,park2025know} have addressed this issue, but each of them focuses only on improving CLIP for a single type of negation task. ~\cite{singh2024learn} introduces CC-Neg, an image-to-text retrieval benchmark built from image-caption-negated caption triplets to evaluate whether CLIP can recognize the negation of objects that are actually present in an image. Based on this benchmark, CoN-CLIP was proposed, which fine-tunes the transformer layers of CLIP and encourages image embeddings to be less aligned with the corresponding negated text embeddings. ~\cite{park2025know} proposes NegRefCOCOg, a text-to-image retrieval benchmark constructed from negated caption-positive image-negative image triplets to evaluate whether CLIP can understand the negation of objects, actions, and attributes that are absent from an image. It further presents NegationCLIP, which fine-tunes transformer layers~\cite{vaswani2017attention} of CLIP, but instead aims to strengthen the alignment between image embeddings and negated text embeddings.

To address the above limitation that existing methods focus only on a single type of negation task, we first formally define two common types of negation tasks: presence-based negation and absence-based negation. Specifically, presence-based negation measures CLIP’s ability to understand negated expressions of objects that are actually present in an image, whereas absence-based negation measures CLIP’s ability to understand negated expressions of objects that may plausibly exist in an image but are in fact absent. Based on these definitions, we propose \textbf{Omni-NegCLIP}, a fine-tuned CLIP model that improves the understanding of both types of negation tasks by adapting the front transformer layers of the CLIP text encoder with our designed presence-based contrastive objective and absence-based contrastive objective. Specifically, the presence-based contrastive objective encourages the model to align each image embedding more closely with its corresponding original caption embedding while explicitly reducing its similarity to the corresponding presence-based negated caption embedding, thereby improving the model’s ability to distinguish an image from negated descriptions of objects that are present. In contrast, the absence-based contrastive objective encourages the model to align each image embedding with both its original caption embedding and its corresponding absence-based negated caption embedding, while explicitly enforcing a semantic distinction between the original caption and the absence-based negated caption in the text embedding space. To evaluate the effectiveness of \textbf{Omni-NegCLIP} in understanding both types of negation tasks, we conduct experiments on different CLIP configurations, including ViT-B/32, ViT-B/16, and ViT-L/14. We evaluate these models on CC-Neg~\cite{singh2024learn} and NegRefCOCOg~\cite{park2025know}, which measure the model's understanding of presence-based negation and absence-based negation, respectively. We also assess the ability of \textbf{Omni-NegCLIP} to preserve CLIP's general capability using the COCO retrieval benchmark~\cite{lin2014microsoft}. Experimental results show that \textbf{Omni-NegCLIP} substantially improves CLIP's understanding of both types of negation tasks, without sacrificing, and in some cases even improving, its general capability in image-text retrieval. Compared with prior methods, \textbf{Omni-NegCLIP} demonstrates stronger ability in understanding multiple types of negation tasks.

Our main contributions are summarized as follows: 1) We propose \textbf{Omni-NegCLIP}, a fine-tuned CLIP model that substantially improves CLIP's ability to understand multiple types of negation tasks, without sacrificing and in some cases even improving its general capability.
2) We introduce the presence-based contrastive objective and the absence-based contrastive objective, which extend the original InfoNCE contrastive loss of CLIP and enhance the model's understanding of presence-based negation and absence-based negation, respectively.
3) Extensive experiments on various CLIP configurations show that, compared with the original pretrained CLIP, our \textbf{Omni-NegCLIP} improves performance by an average of 51.01\% on presence-based negation tasks, 11.54\% on absence-based negation tasks, and 17.24\% on the general COCO retrieval benchmark. Compared with prior methods, \textbf{Omni-NegCLIP} achieves average improvements of 45.26\% on presence-based negation tasks and 17.07\% on absence-based negation tasks, demonstrating its superior effectiveness in understanding multiple types of negation tasks.

\section{Related Work}
\label{sec:Related_work}
\subsection{Vision-Language Models and Limitations}
Recently, VLMs~\cite{li2022blip,alayrac2022flamingo,radford2021learning,singh2022flava,wang2022omnivl,zhai2022lit,jia2021scaling,li2021align,yuan2021florence} have demonstrated strong capabilities in leveraging both visual and textual information to perform a wide range of tasks, including text-to-image (T2I) generation~\cite{rombach2022high}, image-text matching~\cite{li2022blip}, image retrieval~\cite{hendriksen2022extending}, and object classification~\cite{radford2021learning}. Among existing VLMs, CLIP~\cite{radford2021learning} is one of the pioneering and most influential models. It achieves strong performance in zero-shot image classification and semantic search, and has been widely adopted as a backbone for image and video generation models~\cite{ramesh2022hierarchical,singer2022make}, as well as vision-question answering systems~\cite{liu2023visual}. 
However, studies~\cite{kamath2024hard,jiang2024comclip} have shown that CLIP faces challenges in compositional image-text matching, which limits its ability to accurately align structured visual and textual semantics. 
\subsection{Understanding Negation}
To improve CLIP’s compositional image-text matching capability, several methods~\cite{hu2025decoupled,patel2024tripletclip,li2025enhancing,huang2024llm2clip} have been proposed to enhance its ability to understand object order, relations, and attributes. However, negation (e.g., no, not, and without), as an important linguistic phenomenon, is pervasive in natural language yet has been largely overlooked in existing research. As a critical component of sentence structure, negation can alter the entire semantic meaning of a sentence. For example, “this is a photo of a table” and “this is not a photo of a table” convey fundamentally different meanings. Prior studies~\cite{singh2024learn,park2025know,wang2022learn} have shown that CLIP struggles to understand negations in text and often incorrectly associates negated descriptions with corresponding images. Nevertheless, training strategies specifically designed to address CLIP’s difficulty in understanding negation remain largely unexplored. Only a few works~\cite{singh2024learn,park2025know} attempt to tackle this issue, but they are limited to specific contexts. ~\cite{singh2024learn} proposed CC-Neg, a image-to-text retrieval benchmark consisting of image--caption--negated caption triplets, designed to evaluate CLIP’s ability to understand the negation of objects that are present in images. They further introduced CoN-CLIP, which leverages contrastive learning by fine-tuning all transformer layers of CLIP to reduce the alignment between image embeddings and their corresponding negated text embeddings. ~\cite{park2025know} proposed NegRefCOCOg, a text-to-image retrieval benchmark consisting of negated caption–positive image–negative image triplets, designed to evaluate CLIP’s ability to understand the negation of objects, actions, and attributes that are absent from images. They further introduced NegationCLIP, which fine-tunes all transformer layers of CLIP to enhance the alignment between image embeddings and negated text embeddings. However, these two approaches are restricted to negation of concepts that are present in images and negation of concepts that are absent from images, respectively. When evaluated on the opposite type of negation task, they exhibit poor performance, even performing worse than the pretrained CLIP, as shown in Table~\ref{tab:negation_general_benchmark}.

\begin{figure}[t]
    \centering
    \includegraphics[width=0.85\linewidth]{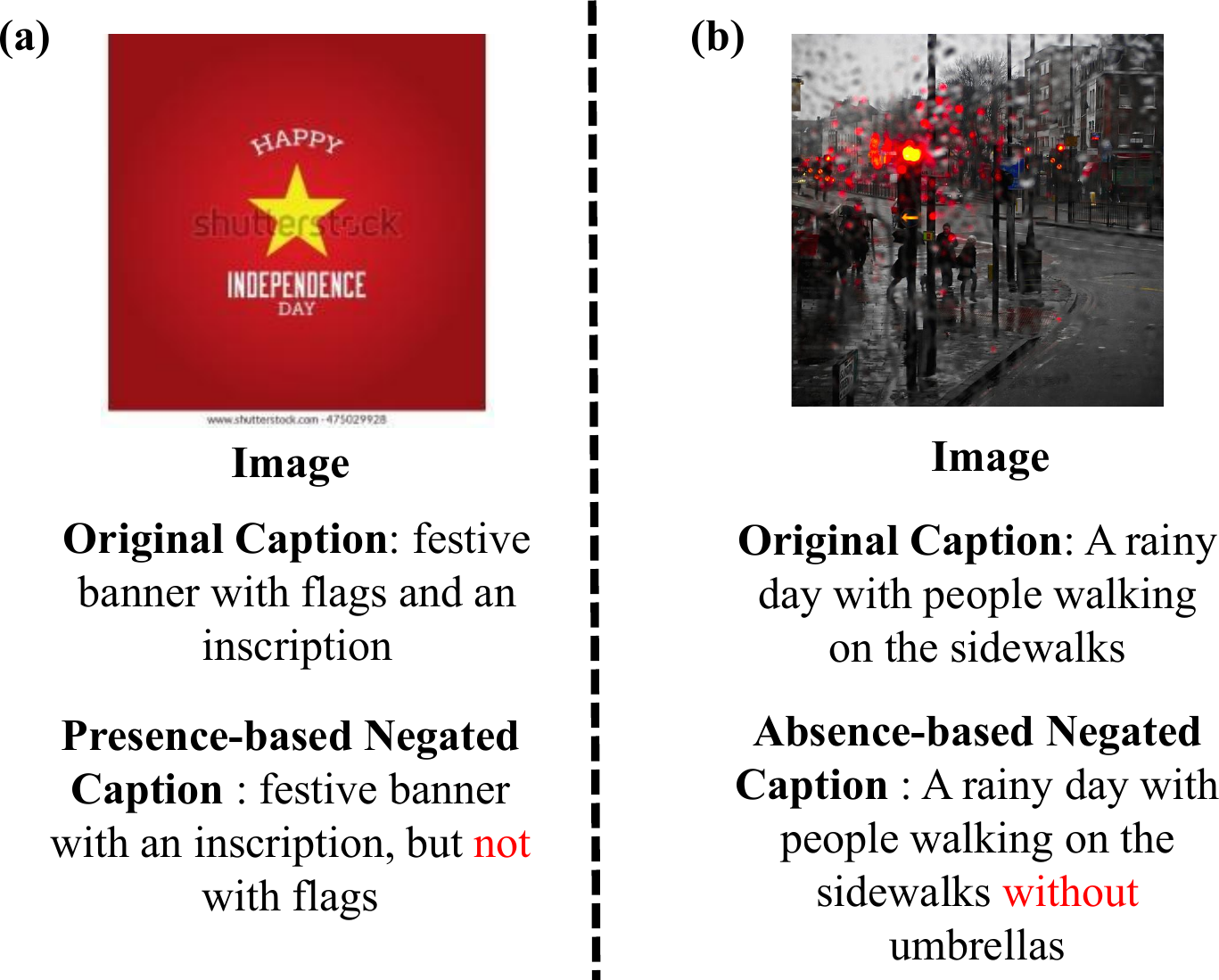 }
    \caption{(a) An example of presence-based negation from CC-Neg~\cite{singh2024learn}. (b) An example of absence-based negation from NegRefCOCOg~\cite{park2025know}.}
    \label{fig:attention_heatmaps}
\end{figure}

\section{Preliminaries}
\label{sec:CLIP Preliminaries}
CLIP~\cite{radford2021learning} is a pretrained model designed to align image semantics with their corresponding caption semantics in a shared embedding space. It consists of two main components: a vision encoder and a text encoder, both of which are composed of multiple transformer layers.
During pretraining, a batch of images, denoted as $\{ I^{(i)} \}_{i=1}^{B}$, and a corresponding batch of captions, denoted as $\{ C^{(i)} \}_{i=1}^{B}$, are processed by the vision encoder and text encoder, respectively, and mapped into a shared latent space. This results in image embeddings $\{ E_I^{(i)} \}_{i=1}^{B}$ and caption embeddings $\{ E_C^{(i)} \}_{i=1}^{B}$.
The model is optimized using a contrastive objective to align each image with its corresponding caption. Specifically, for each image embedding in a batch, the model is trained to pull it closer to its corresponding caption embedding while pushing it away from the other caption embeddings in the batch, using the image-to-caption loss $L_{\text{img}}$. Similarly, for each caption embedding in the batch, the model is trained to pull it closer to its corresponding image embedding while pushing it away from the other image embeddings in the batch, using the caption-to-image loss $L_{\text{txt}}$. The overall contrastive objective $L$ is denoted as:

\begin{equation}
L_{\text{img}}
=
-\frac{1}{B}
\sum_{i=1}^{B}
\log
\frac{
\exp\big(s(E_I^{(i)}, E_C^{(i)})\big)
}{
\sum_{j=1}^{B}
\exp\big(s(E_I^{(i)}, E_C^{(j)})\big)
}
\end{equation}

\begin{equation}
L_{\text{txt}}
=
-\frac{1}{B}
\sum_{i=1}^{B}
\log
\frac{
\exp\big(s(E_C^{(i)}, E_I^{(i)})\big)
}{
\sum_{j=1}^{B}
\exp\big(s(E_C^{(i)}, E_I^{(j)})\big)
}
\end{equation}

\begin{equation}
L
=
\frac{L_{\text{img}} + L_{\text{txt}}}{2}
\end{equation}
where $s(x, y) = \tau \, x^\top y$ denotes the temperature-scaled cosine similarity with learnable temperature parameter $\tau$. After being pretrained to align image semantics with their corresponding caption semantics, CLIP can be directly applied to downstream tasks such as zero-shot image-text matching, image retrieval, and object classification.

\begin{figure*}[t]
    \centering
    \includegraphics[width=1\textwidth]{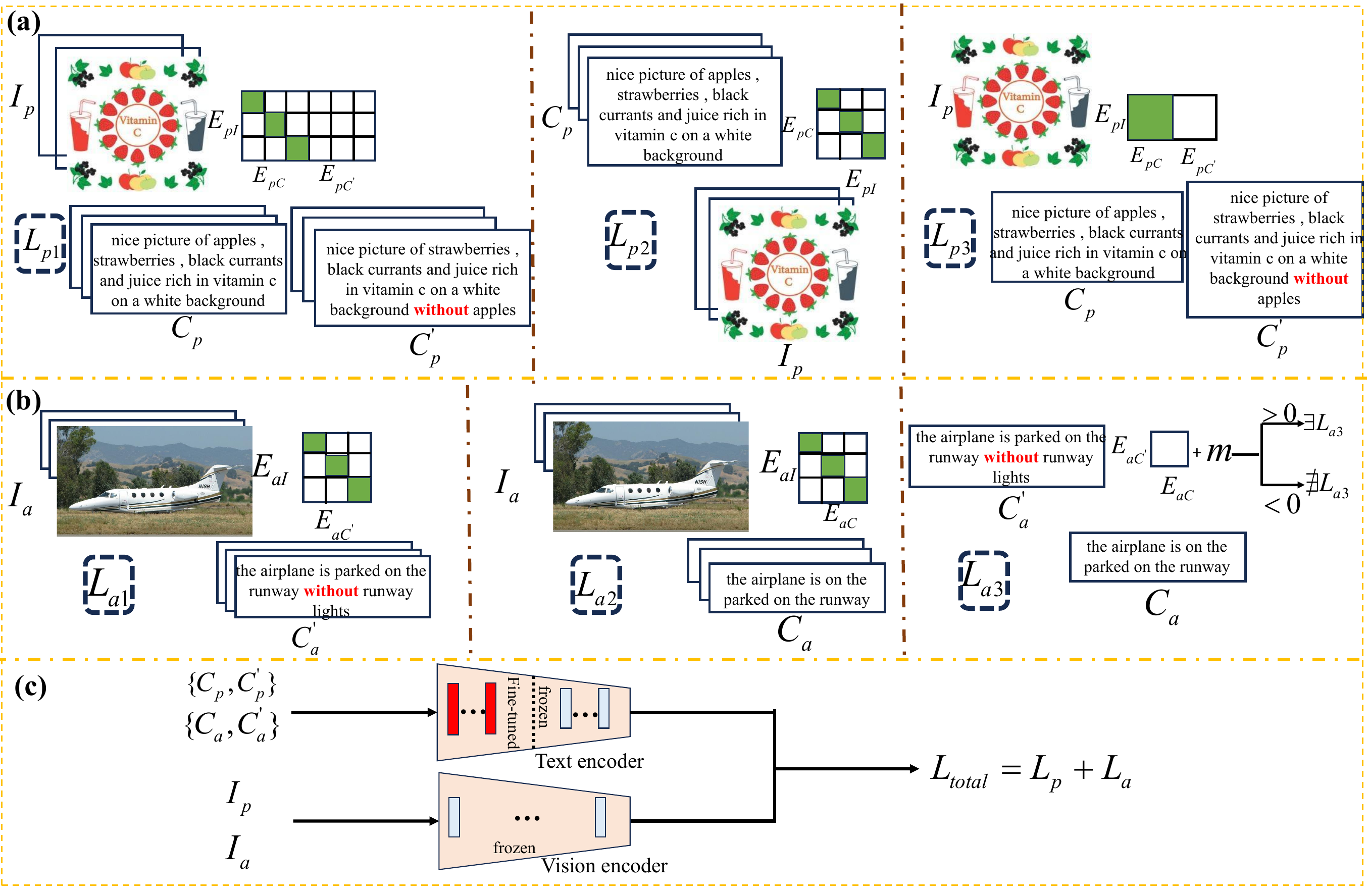}
    \caption{(a) Illustration of our designed presence-based contrastive objective. (b) Illustration of our designed absence-based contrastive objective. (c) Illustration of the fine-tuning pipeline of Omni-NegCLIP.}
    \label{fig:workflow}
\end{figure*}

\section{Proposed Method}
\label{sec:method}
In this section, we first formally define the two types of negation tasks. Then, to enhance CLIP’s understanding of both types of negation, we design appropriate contrastive objectives for fine-tuning based on their respective characteristics. Next, we analyze the differences in learning capacity for negation tasks across transformer layers in the text encoder of CLIP. Finally, we introduce the overall fine-tuning framework of Omni-NegCLIP. Illustrations of the contrastive objectives for both negation types, as well as the fine-tuning framework of Omni-NegCLIP, are shown in Figure~\ref{fig:workflow}.
\subsection{Two Types of Negation Tasks}
\noindent\textbf{Presence-based Negation.}
The presence-based negation task measures CLIP’s ability to understand negated expressions of objects that are present in an image. An example from CC-Neg~\cite{singh2024learn} is shown in Figure~\ref{fig:attention_heatmaps}. We define the construction process of a presence-based negation caption as follows.
Let the original image–caption pair be denoted as $(I_p, C_p)$, where $I_p$ is the image and $C_p$ is its corresponding caption. We decompose the caption $C_p$ into two components:
\begin{equation}
C_p = (r, \{O_1, \dots, O_N\}),
\label{eq:caption_decomposition}
\end{equation}
where $\{O_1, \dots, O_N\}$ denotes the set of object words appearing in $C_p$, and $r$ represents the remaining part of the caption excluding the object words. We randomly sample an object word $O_j \in \{O_1, \dots, O_N\}$ with $j \in \{1, \dots, N\}$, and sample a negation word $\nu \in \{\textit{No}, \textit{Not}, \textit{Without}\}$. We define a negation operator $\mathcal{T}_{\nu}(\cdot)$ that applies the negation word $\nu$ to the selected object word $O_j$, producing a negated object expression:
\begin{equation}
O_j' = \mathcal{T}_{\nu}(O_j),
\label{eq:negated_object}
\end{equation}
The presence-based negated caption corresponding to $I_p$ is constructed by replacing $O_j$ with $O_j'$ in $C_p$, yielding:
\begin{equation}
C_p' = (r, \{O_1, \dots, O_j', \dots, O_N\}).
\label{eq:negated_caption}
\end{equation}

\noindent\textbf{Absence-based Negation.}
The absence-based negation task measures CLIP’s ability to understand negated expressions of objects that may plausibly exist in an image but are in fact absent. An example from NegRefCOCOg~\cite{park2025know} is shown in Figure~\ref{fig:attention_heatmaps}. Specifically, we define the construction process of an absence-based negation caption as follows. Let the original image–caption pair be denoted as $(I_a, C_a)$, where $I_a$ is the image and $C_a$ is its corresponding caption. We decompose the caption $C_a$ into two components:
\begin{equation}
C_a = (r, \{O_1, \dots, O_N\}),
\label{eq:absence_caption_decomposition}
\end{equation}
where $\{O_1, \dots, O_N\}$ denotes the set of object words appearing in $C_a$, and $r$ represents the remaining part of the caption excluding the object words.
We define $O_a$ as an object that may plausibly exist in $I_a$ but is in fact absent. We sample a negation word $\nu \in \{\textit{No}, \textit{Not}, \textit{Without}\}$ and apply the negation operator $\mathcal{T}_{\nu}(\cdot)$ to $O_a$, producing the negated object expression:
\begin{equation}
O_a' = \mathcal{T}_{\nu}(O_a),
\label{eq:absence_negated_object}
\end{equation}
The absence-based negated caption corresponding to $I_a$ is then constructed by adding $O_a'$ to $C_a$, yielding:
\begin{equation}
C_a' = (r, \{O_1, \dots, O_N\}, O_a').
\label{eq:absence_negated_caption}
\end{equation}

As shown in Table~\ref{tab:negation_general_benchmark}, pretrained CLIP models, including CLIP with ViT-B/32, ViT-B/16, and ViT-L/14 backbones, exhibit poor performance on both the presence-based negation benchmark CC-Neg~\cite{singh2024learn} and the absence-based negation benchmark NegRefCOCOg~\cite{park2025know}.

\subsection{Contrastive Objectives}
To enhance the CLIP model’s understanding of the two types of negation tasks, we design separate contrastive objectives for each task. We denote the image encoder as $\mathcal{E}_{img}(\cdot)$, and the text encoder as $\mathcal{E}_{text}(\cdot)$. 

\noindent\textbf{Presence-based Contrastive Objective.}
In the presence-based negation task, the semantics of the image $I_p$ and its corresponding presence-based negated caption $C_p'$ are not aligned. Therefore, the presence-based contrastive objective is designed to push apart the representations of the image and the presence-based negated caption. Specifically, we construct a batch consisting of $B$ triplets, 
denoted as $\mathcal{B}_p = \{(I_p^{(i)}, C_p^{(i)}, C_p^{\prime (i)})\}_{i=1}^{B}$. The images $\{ I_p^{(i)} \}_{i=1}^{B}$ are encoded by $\mathcal{E}_{img}(\cdot)$ to obtain image embeddings $\{ E_{pI}^{(i)} \}_{i=1}^{B}$. 
The captions $\{ C_p^{(i)} \}_{i=1}^{B}$ are encoded by $\mathcal{E}_{text}(\cdot)$ to obtain caption embeddings $\{ E_{pC}^{(i)} \}_{i=1}^{B}$. 
The presence-based negated captions $\{ C_p^{\prime (i)} \}_{i=1}^{B}$ are encoded by $\mathcal{E}_{text}(\cdot)$ to obtain negated caption embeddings $\{ E_{pC'}^{(i)} \}_{i=1}^{B}$.
The presence-based contrastive objective consists of the following three loss terms.

\noindent\textbf{1. Images-to-All-Captions Contrastive Loss:}
This loss term pulls each image embedding closer to its corresponding caption embedding while implicitly pushing it away from the corresponding presence-based negated caption embedding. Specifically, we concatenate $\{ E_{pC}^{(i)} \}_{i=1}^{B}$ and $\{ E_{pC'}^{(i)} \}_{i=1}^{B}$ to form the complete caption embedding set:
\begin{equation}
\mathcal{T}_p =
\left\{
E_{pC}^{(i)}
\right\}_{i=1}^{B}
\cup
\left\{
E_{pC'}^{(i)}
\right\}_{i=1}^{B}.
\label{eq:Tp}
\end{equation}
The image-to-all-captions contrastive loss $\mathcal{L}_{p1}$ is computed as
\begin{equation}
\mathcal{L}_{p1}
=
-\frac{1}{B}
\sum_{i=1}^{B}
\log
\frac{\exp(s(E_{pI}^{(i)}, E_{pC}^{(i)}))}
{\sum_{t \in \mathcal{T}_p} \exp(s(E_{pI}^{(i)}, t))}.
\label{eq:p1}
\end{equation}

\noindent\textbf{2. Captions-to-Images Contrastive Loss:}
In this loss term, we align captions with their corresponding images. The text-to-image contrastive loss $\mathcal{L}_{p2}$ is computed as
\begin{equation}
\mathcal{L}_{p2}
=
-\frac{1}{B}
\sum_{i=1}^{B}
\log
\frac{\exp(s(E_{pC}^{(i)}, E_{pI}^{(i)}))}
{\sum_{j=1}^{B} \exp(s(E_{pC}^{(i)}, E_{pI}^{(j)}))}.
\label{eq:p2}
\end{equation}

\noindent\textbf{3. Explicit Negation Discrimination Loss:}
In this loss term, we explicitly encourage the model to distinguish between the caption and the corresponding presence-based negation caption. Specifically, we pull each image embedding closer to its corresponding caption embedding while pushing it away from the corresponding presence-based negation caption embedding. The explicit negation discrimination loss $\mathcal{L}_{p3}$ is computed as
\begin{equation}
\mathcal{L}_{p3}
=
-\frac{1}{B}
\sum_{i=1}^{B}
\log
\frac{\exp(s(E_{pI}^{(i)}, E_{pC}^{(i)}))}
{\exp(s(E_{pI}^{(i)}, E_{pC}^{(i)}))
+
\exp(s(E_{pI}^{(i)}, E_{pC'}^{(i)}))}.
\label{eq:p3}
\end{equation}

The total presence-based contrastive objective is computed as $\mathcal{L}_{p} = (\mathcal{L}_{p1} + \mathcal{L}_{p2} + \mathcal{L}_{p3}) / 3$. We denote the overall objective construction process as 
$\mathcal{L}_{p} = f_p(\mathcal{B}_p)$, 
where $f_p(\cdot)$ represents the mapping from $\mathcal{B}_p$ to the corresponding objective value.

\noindent\textbf{Absence-based Contrastive Objective.}
In the absence-based negation task, the CLIP model is expected to align the image semantics with both the corresponding caption semantics and the absence-based negation caption semantics, while being able to distinguish the subtle semantic differences between the caption and the absence-based negation caption. Specifically, we construct a batch consisting of $B$ triplets, denoted as 
$\mathcal{B}_a = \{(I_a^{(i)}, C_a^{(i)}, C_a^{\prime (i)})\}_{i=1}^{B}$. 
Using $\mathcal{E}_{img}(\cdot)$ and $\mathcal{E}_{text}(\cdot)$, we obtain the image embeddings, caption embeddings, and absence-based negation caption embeddings as 
$\{ E_{aI}^{(i)} \}_{i=1}^{B}$, 
$\{ E_{aC}^{(i)} \}_{i=1}^{B}$, 
and 
$\{ E_{aC'}^{(i)} \}_{i=1}^{B}$, respectively.

\noindent\textbf{1. Image-to-Negation Contrastive Loss:} The first loss aligns images with their corresponding absence-based negation captions. The loss $\mathcal{L}_{a1}$ is defined as
\begin{equation}
\begin{aligned}
\mathcal{L}_{a1}
=
-\frac{1}{2B}
\sum_{i=1}^{B}
\Bigg[
&\log
\frac{\exp(s(E_{aI}^{(i)}, E_{aC'}^{(i)}))}
{\sum_{j=1}^{B}\exp(s(E_{aI}^{(i)}, E_{aC'}^{(j)}))}
\\
&+
\log
\frac{\exp(s(E_{aC'}^{(i)}, E_{aI}^{(i)}))}
{\sum_{j=1}^{B}\exp(s(E_{aC'}^{(i)}, E_{aI}^{(j)}))}
\Bigg].
\end{aligned}
\label{eq:a1}
\end{equation}

\noindent\textbf{2. Image-to-Caption Contrastive Loss:}
To preserve the model’s ability to align image semantics with caption semantics, we define
\begin{equation}
\begin{aligned}
\mathcal{L}_{a2}
=
-\frac{1}{2B}
\sum_{i=1}^{B}
\Bigg[
&\log
\frac{\exp(s(E_{aI}^{(i)}, E_{aC}^{(i)}))}
{\sum_{j=1}^{B}\exp(s(E_{aI}^{(i)}, E_{aC}^{(j)}))}
\\
&+
\log
\frac{\exp(s(E_{aC}^{(i)}, E_{aI}^{(i)}))}
{\sum_{j=1}^{B}\exp(s(E_{aC}^{(i)}, E_{aI}^{(j)}))}
\Bigg].
\end{aligned}
\label{eq:a2}
\end{equation}

\noindent\textbf{3. Caption Contrastive Loss:}
To explicitly distinguish the subtle semantic differences between the original caption and its corresponding absence-based negation caption in the text embedding space, we introduce a penalty term whenever their similarity exceeds a margin of $-m$. Specifically, we define
\begin{equation}
\mathcal{L}_{a3}
=
\frac{1}{B}
\sum_{i=1}^{B}
\max\left(
0,\,
s(E_{aC}^{(i)}, E_{aC'}^{(i)}) + m
\right),
\label{eq:a3}
\end{equation}

The total absence-based contrastive objective is computed as $\mathcal{L}_{a} = (\mathcal{L}_{a1} + \mathcal{L}_{a2} + \mathcal{L}_{a3}) / 3$. We denote the overall objective construction process as 
$\mathcal{L}_{a} = f_a(\mathcal{B}_a)$, 
where $f_a(\cdot)$ represents the mapping from $\mathcal{B}_a$ to the corresponding objective value.

\begin{figure}[t]
    \centering
    \includegraphics[width=0.9\linewidth]{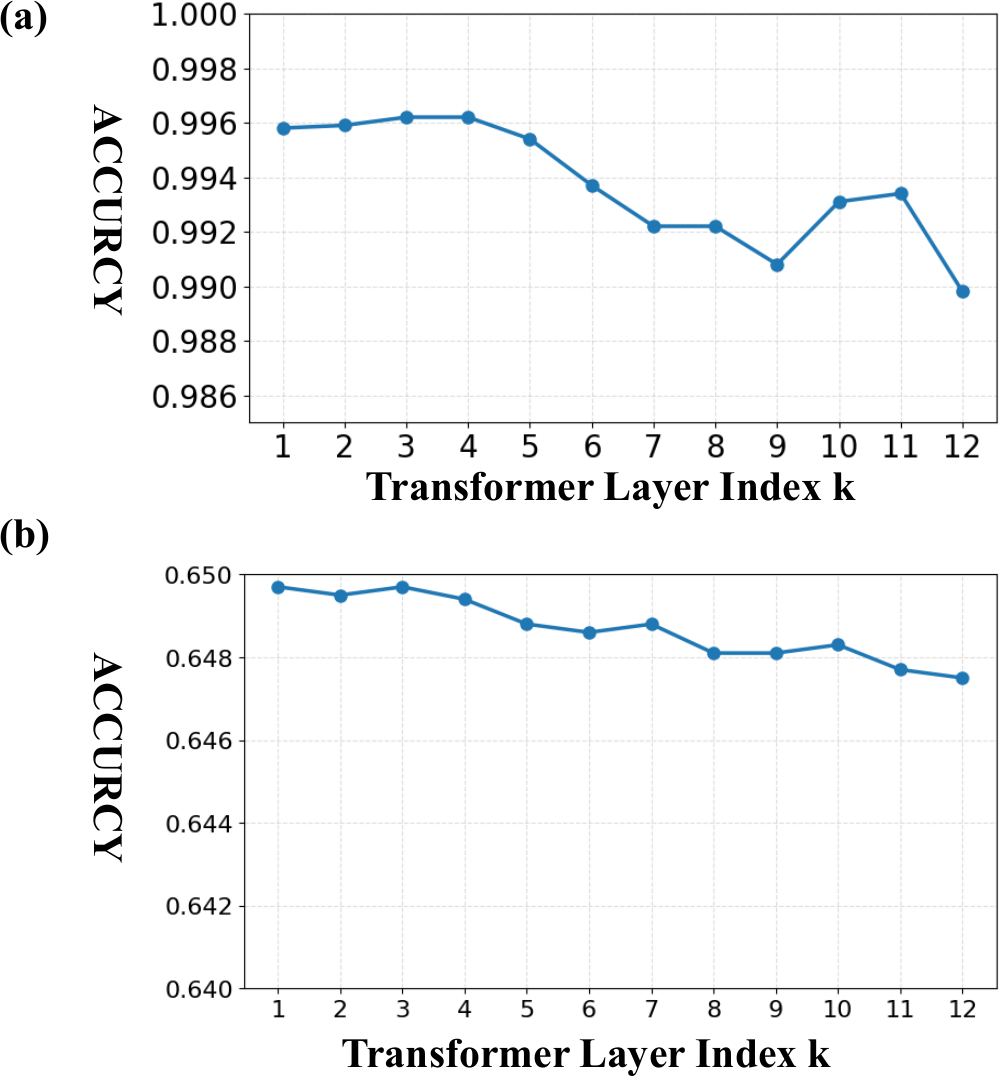}
    \caption{(a) presence-based negation learning capability across transformer layers in the CLIP text encoder on CC-Neg~\cite{singh2024learn}. (b) absence-based negation learning capability across transformer layers in the CLIP text encoder on NegRefCOCOg~\cite{park2025know}. Front layers show higher accuracy than later layers, indicating stronger negation learning capability.}
    \label{fig:layer wise accuracy}
\end{figure}

\subsection{Layer-wise Negation Learning Analysis}
\label{sec:layer}
Using the proposed presence-based contrastive objective $\mathcal{L}_{p}$ and absence-based contrastive objective $\mathcal{L}_{a}$, we explore the differences in negation learning capability across the transformer layers of the CLIP text encoder.
Specifically, we denote the transformer layers in the text encoder as $\{ \ell_k \}_{k=1}^{12}$. For each layer $\ell_k$, we fine-tune only $\ell_k$ and the projection layer $\mathcal{P}(\cdot)$ of a pretrained CLIP ViT-B/32 model, while keeping all other parameters frozen.
For each $\ell_k$, we conduct two independent experiments: one fine-tuned using $\mathcal{L}_{p}$ and evaluated on the presence-based benchmark, and the other fine-tuned using $\mathcal{L}_{a}$ and evaluated on the absence-based benchmark. All experiments follow the same setup as described in Section~\ref{sec:Experiment}, and the results are shown in Figure~\ref{fig:layer wise accuracy}. As can be observed, for both types of negation tasks, fine-tuning the front transformer layers achieves significantly higher accuracy than fine-tuning the later transformer layers. This indicates that the front transformer layers possess stronger negation learning capability.
We hypothesize that this is because understanding negation in text requires the CLIP text encoder to capture fine-grained syntactic information in the sentence, specifically identifying the scope or target of the negation. According to~\cite{dumpala2024seeing}, the front layers of the CLIP text encoder exhibit stronger capabilities in modeling such fine-grained syntactic structures compared to the later layers.

\subsection{Omni-NegCLIP Fine-tuning Pipeline}
To enable the final Omni-NegCLIP model to simultaneously understand both types of negation tasks, we optimize the model using a combined objective in each fine-tuning step, where the total loss is defined as the sum of $\mathcal{L}_{p}$ and $\mathcal{L}_{a}$.
Furthermore, as revealed in Section~\ref{sec:layer}, the front transformer layers exhibit stronger negation learning capability than the later layers. Therefore, we fine-tune only the front transformer layers $\mathcal{F} = \{\ell_k\}_{k=1}^{K}$ and the projection layer $\mathcal{P}(\cdot)$, while keeping all other parameters frozen.
The detailed procedure of one fine-tuning step is described in Algorithm~\ref{alg:omni_negclip}.

\begin{algorithm}[t]
\LinesNumbered
\caption{Omni-NegCLIP Fine-tuning}
\label{alg:omni_negclip}
\KwIn{
Pretrained text encoder $\mathcal{E}_{text}(\cdot)$; 
Front transformer layers $\mathcal{F} = \{\ell_k\}_{k=1}^{K}$; 
Projection layer $\mathcal{P}(\cdot)$; 
Presence-based batch $\mathcal{B}_p = \{(I_p^{(i)}, C_p^{(i)}, C_p'^{(i)})\}_{i=1}^{B}$; 
Absence-based batch $\mathcal{B}_a = \{(I_a^{(i)}, C_a^{(i)}, C_a'^{(i)})\}_{i=1}^{B}$
}
\KwOut{
Updated $\mathcal{E}_{text}(\cdot)$
}
\SetAlgoLined

\textit{/* Compute losses in one fine-tuning step */}\\
$\mathcal{L}_p \gets f_p(\mathcal{B}_p)$\;
$\mathcal{L}_a \gets f_a(\mathcal{B}_a)$\;
$\mathcal{L}_{total} \gets \mathcal{L}_p + \mathcal{L}_a$\;

\textit{/* Update parameters */}\\
$\mathcal{E}_{text}(\cdot) \leftarrow 
\text{update}\!\left(
\mathcal{E}_{text}(\cdot), 
\mathcal{L}_{total}, 
\mathcal{F}, 
\mathcal{P}(\cdot)
\right)$\;

\Return{$\mathcal{E}_{text}(\cdot)$}
\end{algorithm}

\section{Experimental Results}
\label{sec:Experiment}
In this section, we systematically evaluate the capability of Omni-NegCLIP in understanding both presence-based negation and absence-based negation tasks, as well as its ability to retain general capability. We conduct experiments based on three CLIP configurations: ViT-B/32, ViT-B/16, and ViT-L/14.

\subsection{Experimental Setup}
\label{sec: setup}
During fine-tuning of Omni-NegCLIP, we use object-absence negation (OAN) data~\cite{singh2024learn} as absence-based samples and 188,246 triplets from CC-Neg~\cite{singh2024learn} as presence-based samples. The batch size is set to $B = 128$, and the model is trained for 30 epochs. The maximum index of the front transformer layers is set to $K = 6$. We set the margin parameter to $m = 0.9$ in the caption contrastive loss. The learning rate is set to 1e-6, and the model is optimized using the AdamW optimizer.
Under each CLIP configuration, we compare Omni-NegCLIP with the pretrained CLIP and two prior fine-tuned negation comprehension models, CoN-CLIP~\cite{singh2024learn} and NegationCLIP~\cite{park2025know}, on two negation comprehension benchmarks: NegRefCOCOg~\cite{park2025know} and CC-Neg~\cite{singh2024learn}. NegRefCOCOg is used to evaluate absence-based negation understanding, while CC-Neg is used to evaluate presence-based negation understanding. Please note that for CC-Neg evaluation, we use the portion of the dataset excluding the 188,246 samples used for fine-tuning.
Furthermore, we evaluate text-to-image retrieval performance on the COCO retrieval benchmar~\cite{lin2014microsoft} to assess whether general capability is preserved after fine-tuning. For the original CLIP and the two prior methods, we use their publicly released checkpoints. All experiments are conducted on NVIDIA V100 GPUs.








\begin{table}[t]
\centering
\scriptsize
\setlength{\tabcolsep}{4pt}
\renewcommand{\arraystretch}{1.3}
\resizebox{\columnwidth}{!}{
\begin{tabular}{l c c c c c}
\toprule
\textbf{Model} & \textbf{Arch.} & \textbf{CC-Neg~\cite{singh2024learn}} & \textbf{NegRefCOCOg~\cite{park2025know}} & \textbf{Avg.} & \textbf{COCO~\cite{lin2014microsoft}} \\
\midrule

CLIP & \multirow{4}{*}{ViT-B/32} & 65.70 & 57.73 & 61.72 & 40.88 \\
CoN-CLIP~\cite{singh2024learn} &  & \textbf{99.70} & 55.23 & 77.47 & 41.72 \\
NegationCLIP~\cite{park2025know} &  & 66.20 & 63.86 & 65.03 & \textbf{52.12} \\
Omni-NegCLIP &  & 99.60 & \textbf{64.55} & \textbf{82.08} & 48.90 \\

\midrule

CLIP & \multirow{4}{*}{ViT-B/16} & 65.22 & 58.18 & 61.70 & 42.90 \\
CoN-CLIP~\cite{singh2024learn} &  & \textbf{99.70} & 55.45 & 77.58 & 45.38 \\
NegationCLIP~\cite{park2025know} &  & 69.11 & 65.00 & 67.06 & \textbf{53.86} \\
Omni-NegCLIP &  & 99.56 & \textbf{65.45} & \textbf{82.51} & 51.06 \\

\midrule

CLIP & \multirow{4}{*}{ViT-L/14} & 66.89 & 57.27 & 62.08 & 47.61 \\
CoN-CLIP~\cite{singh2024learn} &  & \textbf{99.69} & 55.68 & 77.69 & 47.90 \\
NegationCLIP~\cite{park2025know} &  & 70.30 & 62.50 & 66.40 & \textbf{56.78} \\
Omni-NegCLIP &  & 99.52 & \textbf{63.18} & \textbf{81.35} & 53.84 \\

\bottomrule
\end{tabular}
}
\caption{Comparison of the performance of different models on negation tasks and the general task across different CLIP architectures. Avg. denotes the average accuracy on CC-Neg~\cite{singh2024learn} and NegRefCOCOg~\cite{park2025know}. Bold numbers indicate the best performance within each architecture group.}
\label{tab:negation_general_benchmark}
\end{table}

\begin{table}[t]
\centering
\renewcommand{\arraystretch}{1.1}
{\fontsize{7pt}{8.5pt}\selectfont
\setlength{\tabcolsep}{1.5pt}
\begin{tabularx}{\linewidth}{@{}lcc@{}}
\toprule
\textbf{Choice} & \textbf{CC-Neg}~\cite{singh2024learn} & \textbf{NegRefCOCOg}~\cite{park2025know} \\
\midrule
\multicolumn{3}{l}{\textbf{Influence of Loss Terms}} \\
\multicolumn{3}{l}{$(\{\ell_k\}_{k=1}^{6},\; m = 0.9)$} \\
\quad Omni-NegCLIP(w/o $\mathcal{L}_{p3}$) & 99.44 & 63.41 \\
\quad Omni-NegCLIP(w/o $\mathcal{L}_{a1}$) & 99.63 & 54.77 \\
\quad Omni-NegCLIP(w/o $\mathcal{L}_{a3}$) & \textbf{99.66} & 61.82 \\
\quad \textbf{Omni-NegCLIP(all loss terms)} & 99.60 & \textbf{64.55} \\
\midrule
\multicolumn{3}{l}{\textbf{Influence of Layers}} \\
\multicolumn{3}{l}{$(m = 0.9)$} \\
\quad Omni-NegCLIP($\{\ell_k\}_{k=1}^{4}$) & 99.56 & 62.95 \\
\quad Omni-NegCLIP($\{\ell_k\}_{k=1}^{5}$) & 99.58 & 62.73 \\
\quad \textbf{Omni-NegCLIP($\{\ell_k\}_{k=1}^{6}$)} & \textbf{99.60} & \textbf{64.55} \\
\quad Omni-NegCLIP($\{\ell_k\}_{k=1}^{12}$) & 99.47 & 63.18 \\
\midrule
\multicolumn{3}{l}{\textbf{Influence of Margin Value $m$}} \\
\multicolumn{3}{l}{$(\{\ell_k\}_{k=1}^{6})$} \\
\quad Omni-NegCLIP($(m = 0)$) & \textbf{99.70} & 59.77 \\
\quad Omni-NegCLIP($(m = 0.1)$) & 99.69 & 60.68 \\
\quad Omni-NegCLIP($(m = 0.2)$) & 99.69 & 59.55 \\
\quad Omni-NegCLIP($(m = 0.3)$) & 99.68 & 60.23 \\
\quad Omni-NegCLIP($(m = 0.4)$) & 99.68 & 60.45 \\
\quad Omni-NegCLIP($(m = 0.5)$) & 99.69 & 61.82 \\
\quad Omni-NegCLIP($(m = 0.6)$) & 99.61 & 62.27 \\
\quad Omni-NegCLIP($(m = 0.7)$) & 99.60 & 62.73 \\
\quad Omni-NegCLIP($(m = 0.8)$) & 99.53 & 63.18 \\
\quad \textbf{Omni-NegCLIP($(m = 0.9)$)} & 99.60 & \textbf{64.55} \\
\bottomrule
\end{tabularx}
}
\caption{Ablation study on loss terms, the selection of transformer layers in the text encoder during fine-tuning, and the margin value $m$ in the caption contrastive loss $\mathcal{L}_{a3}$.}
\label{tab:ablation}
\end{table}

\subsection{Main Results}
\label{sec: results}
In Table~\ref{tab:negation_general_benchmark}, we present the accuracy results of these models on CC-Neg, NegRefCOCOg, and the COCO retrieval benchmark across CLIP ViT-B/32, CLIP ViT-B/16, and CLIP ViT-L/14 configurations. As can be observed, compared with the pretrained CLIP, Omni-NegCLIP achieves significant improvements on all three configurations. On CC-Neg, the accuracy improvements are 51.60\%, 52.65\%, and 48.78\% for ViT-B/32, ViT-B/16, and ViT-L/14, respectively. On NegRefCOCOg, the improvements are 11.81\%, 12.50\%, and 10.32\%, respectively. On the COCO retrieval benchmark, Omni-NegCLIP improves accuracy by 19.62\%, 19.02\%, and 13.09\%, respectively.
Compared with CoN-CLIP, on CC-Neg, Omni-NegCLIP achieves slightly lower accuracy by an average of 0.13\% across the three CLIP configurations. However, we emphasize that the accuracy of Omni-NegCLIP consistently exceeds 99\%, which is already sufficient to demonstrate strong presence-based negation understanding capability. On NegRefCOCOg, Omni-NegCLIP outperforms CoN-CLIP by 16.87\%, 18.03\%, and 13.47\% across ViT-B/32, ViT-B/16, and ViT-L/14, respectively. On COCO retrieval, Omni-NegCLIP achieves improvements of 17.21\%, 12.52\%, and 12.40\%, respectively.
Compared with NegationCLIP, Omni-NegCLIP achieves higher accuracy on both CC-Neg and NegRefCOCOg across all three CLIP configurations. Specifically, on CC-Neg, our method improves accuracy by an average of 45.36\%, and on NegRefCOCOg, the average improvement is 0.96\%. Although on the COCO retrieval benchmark Omni-NegCLIP performs 5.51\% lower on average than NegationCLIP, we emphasize that it still achieves higher accuracy than the pretrained CLIP. This demonstrates that our method is able to maintain, and even improve, the general capability of the model after fine-tuning.

\noindent\textbf{Discussion.} Overall, compared with the pretrained CLIP, Omni-NegCLIP achieves significantly higher accuracy on both negation tasks. Moreover, rather than suffering performance degradation on the general task of COCO retrieval, it even improves performance to a certain extent. 
Compared with CoN-CLIP and NegationCLIP, which each perform well on only a single type of negation task, our method demonstrates a more comprehensive understanding of negation.

\section{Ablation}
In this study, we evaluate the effectiveness of key components in our Omni-NegCLIP, including the loss terms in the contrastive objectives, the selection of front transformer layers, and the margin value $m$ used in the caption contrastive loss. The results are shown in Table~\ref{tab:ablation}. 
For fairness, all experiments are conducted on CLIP ViT-B/32 using the CC-Neg and NegRefCOCOg benchmarks.

\noindent\textbf{Influence of Loss Terms.} 
We evaluate the impact of different loss terms in the presence-based and absence-based contrastive objectives on model performance. As shown in Table~\ref{tab:ablation}, Omni-NegCLIP (w/o $\mathcal{L}_{p3}$) refers to the variant without the explicit negation discrimination loss $\mathcal{L}_{p3}$ in the presence-based contrastive objective. It performs worse than the full Omni-NegCLIP on the CC-Neg benchmark. This may be because, without explicitly encouraging the model to push the image embedding away from the corresponding presence-based negation caption embedding, the model’s understanding of presence-based negation is weakened.
Furthermore, Omni-NegCLIP (w/o $\mathcal{L}_{a1}$) denotes the variant without the image-to-negation contrastive loss $\mathcal{L}_{a1}$ in the absence-based contrastive objective. Its accuracy on the NegRefCOCOg benchmark is 13.63\% lower than that of the full Omni-NegCLIP. This may be because, without aligning the image representation with the corresponding absence-based negation caption, the model struggles to properly learn the absence-based negation task.
However, merely aligning the image with the absence-based negation caption is insufficient. Omni-NegCLIP (w/o $\mathcal{L}_{a3}$) removes the caption-level contrastive loss $\mathcal{L}_{a3}$ and achieves 2.51\% lower accuracy on NegRefCOCOg compared to the full model. This may be because, under $\mathcal{L}_{a1}$ and $\mathcal{L}_{a2}$, the model aligns the image representation with both the absence-based negation caption and the original caption, which may cause semantic confusion between the two. The loss $\mathcal{L}_{a3}$ explicitly enforces separation between the original caption and the absence-based negation caption in the text embedding space.

\noindent\textbf{Influence of Layers.} 
We evaluate the impact of selecting different numbers of front transformer layers in Omni-NegCLIP on model performance. As shown in Table~\ref{tab:ablation}, using the first four transformer layers, denoted as Omni-NegCLIP($\{\ell_k\}_{k=1}^{4}$), yields lower accuracy on both CC-Neg and NegRefCOCOg compared to using the first six layers, denoted as Omni-NegCLIP($\{\ell_k\}_{k=1}^{6}$).
We hypothesize that although fine-tuning only the first four transformer layers achieves better performance than fine-tuning later layers when the model is trained to understand each negation task independently, as shown in Figure~\ref{fig:layer wise accuracy}, jointly learning both presence-based and absence-based negation tasks requires sufficient model capacity to capture knowledge from both tasks. Therefore, relying solely on the parameters of the first four layers may be insufficient.
However, using all transformer layers, i.e., Omni-NegCLIP($\{\ell_k\}_{k=1}^{12}$), results in lower accuracy on both CC-Neg and NegRefCOCOg compared to Omni-NegCLIP($\{\ell_k\}_{k=1}^{6}$). This suggests that once sufficient capacity is provided to learn both negation types, further increasing the number of trainable layers does not enhance negation understanding. Instead, the relatively weaker capability of the later transformer layers in modeling fine-grained syntactic structures in captions may hinder the model’s ability to effectively learn negation.

\noindent\textbf{Influence of Margin Value.} 
We evaluate the impact of the margin value $m$ in the caption contrastive loss $\mathcal{L}_{a3}$ of the absence-based contrastive objective on model performance. As shown in Table~\ref{tab:ablation}, Omni-NegCLIP with $m=0.9$ achieves significantly higher accuracy on the NegRefCOCOg benchmark compared to other margin settings.
We hypothesize that this is because a smaller margin value is insufficient to enforce adequate separation between the original caption and the absence-based negation caption in the embedding space, resulting in semantic ambiguity.

\section{Conclusion}
In this work, we propose \textbf{Omni-NegCLIP}, a fine-tuned CLIP model that improves the understanding of two common types of negation tasks, namely presence-based negation and absence-based negation, by fine-tuning the front transformer layers of CLIP's text encoder with our designed contrastive objectives. This design is motivated by our observation that these front transformer layers exhibit stronger learning ability for negated text than the later layers. Specifically, we design a presence-based contrastive objective that pulls image embeddings closer to their original caption embeddings while pushing them away from the corresponding presence-based negated caption embeddings, and an absence-based contrastive objective that aligns image embeddings with both original and absence-based negated caption embeddings while maintaining a semantic distinction between the two text embeddings. Extensive evaluations on different CLIP configurations demonstrate that \textbf{Omni-NegCLIP} consistently improves the understanding of both types of negation tasks, while preserving and even improving the model's general capability on image-text retrieval.
\label{sec:Methodology}

{
    \small
    \bibliographystyle{ieeenat_fullname}
    \bibliography{main}
}


\end{document}